\documentclass[conference]{IEEEtran}
\IEEEoverridecommandlockouts

\usepackage{amsmath}
\usepackage[table]{xcolor} 
\usepackage[numbers,sort&compress]{natbib}
\usepackage{nicefrac}
\usepackage{graphicx}
\usepackage{subcaption} % added for subfigures
\usepackage{multirow}
\usepackage{comment}
\usepackage{soul}
\usepackage{xcolor}
\usepackage[none]{hyphenat}
\usepackage{url}
\usepackage{tikz}
\usepackage{pgfplots}
\pgfplotsset{compat=1.18}
\usepackage{environ}
\usetikzlibrary{calc}

\def\BibTeX{{\rm B\kern-.05em{\sc i\kern-.025em b}\kern-.08em
    T\kern-.1667em\lower.7ex\hbox{E}\kern-.125emX}}
    
\begin{document}

\title{Contrastive Self-Supervised Learning at the Edge: \\ An Energy Perspective\\
\thanks{This work has been funded by the grant CHIST-ERA-20-SICT-004 (SONATA) by PCI2021-122043-2A/AEI/10.13039/501100011033 and by the European Union’s Horizon
2020 Marie Skłodowska Curie Innovative Training Network Greenedge (GA.
No. 953775). This work has also been partially supported by the ``Ministerio de Asuntos Económicos y Transformación Digital" and the European Union-NextGenerationEU in the frameworks of the ``Plan de Recuperación, Transformación y Resiliencia" and of the ``Mecanismo de Recuperación y Resiliencia" under project SUCCESS-6G with reference TSI-063000-2021-39, TSI-063000-2021-40, and TSI-063000-2021-41. The work was partially done while Fernanda Famá was visiting Nokia Bell Labs.}
%\vspace{-1em}
}

\author{\IEEEauthorblockN{Fernanda Famá\IEEEauthorrefmark{1}, Roberto Pereira\IEEEauthorrefmark{1}, Charalampos Kalalas\IEEEauthorrefmark{1}, Paolo Dini\IEEEauthorrefmark{1}, \\ Lorena Qendro\IEEEauthorrefmark{3}, Fahim Kawsar\IEEEauthorrefmark{2}\IEEEauthorrefmark{3}, Mohammad Malekzadeh\IEEEauthorrefmark{3}}
\IEEEauthorblockA{{\IEEEauthorrefmark{1}Centre Tecnològic de Telecomunicacions de Catalunya (CTTC/CERCA)}
\IEEEauthorblockA{\IEEEauthorrefmark{2}{University of Glasgow, UK}\IEEEauthorrefmark{3}{Nokia Bell Labs, Cambridge, UK}}
Email: \{fernanda.fama, rpereira, ckalalas, pdini\}@cttc.es,} 
\{lorena.qendro, fahim.kawsar, mohammad.malekzadeh\}@nokia-bell-labs.com}

\maketitle

\begin{abstract}
While contrastive learning (CL) shows considerable promise in self-supervised representation learning, its deployment on resource-constrained devices remains largely underexplored. 
The substantial computational demands required for training conventional CL frameworks pose a set of challenges, particularly in terms of energy consumption, data availability, and memory usage. 
We conduct an evaluation of four widely used CL frameworks: SimCLR, MoCo, SimSiam, and Barlow Twins. We focus on the practical feasibility of these CL frameworks for edge and fog deployment, and introduce a systematic benchmarking strategy that includes energy profiling and reduced training data conditions. Our findings reveal that SimCLR, contrary to its perceived computational cost, demonstrates the lowest energy consumption across various data regimes. Finally, we also extend our analysis by evaluating lightweight neural architectures when paired with CL frameworks. 
Our study aims to provide insights into the resource implications of deploying CL in edge/fog environments with limited processing capabilities and opens several research directions for its future optimization. 

\end{abstract}


%\vspace{-0.1cm}
\section{Introduction }
\label{sec:intro}

Over the years, a variety of contrastive learning (CL) approaches have been developed, including popular frameworks such as SimCLR~\citep{chen2020simple}, MoCo~\citep{he2020momentum}, BYOL~\citep{grill2020bootstrap}, SimSiam~\citep{chen2021exploring}, and Barlow Twins~\citep{zbontar2021barlow}, each offering specific advantages and trade-offs. These frameworks aim to learn representations by distinguishing between similar (positive) and dissimilar (negative) samples in a latent space. While some methods rely on large negative sample sets to achieve high-quality representations, others bypass the need for negative pairs through momentum encoders or predictor networks. 
Despite the remarkable advances in CL, deploying these methods on resource-constrained devices remains challenging. A major limitation is the substantial amount of training data required by most CL frameworks to learn meaningful representations \cite{Shi_Zhang_Tang_Zhu_Li_Guo_Zhuang_2022}. Moreover, training typically involves repeated augmentation of data samples, which significantly increases the memory footprint and computational overhead. Many state-of-the-art CL approaches also depend on large batch sizes or memory banks to store negative samples, making them heavily reliant on powerful processing units, often unavailable in edge/fog environments \cite{gui_survey_2024}. Furthermore, optimizing neural networks (NNs) for resource-limited devices is challenging, especially considering the widespread adoption of the computationally intensive ResNet-50 as the standard backbone in CL studies~\cite{Lin_Ding_Cao_Zheng_2023}.

Acquiring sufficiently large datasets for training is another major obstacle in many real-world applications, particularly in distributed setups (e.g., cyber-physical systems, Internet of Things (IoT)) and privacy-sensitive domains such as healthcare and finance \cite{cao_improving_2024}. Unlike in data center settings, where massive datasets can be collected and processed on high-performance servers, edge computing relies on resource-limited devices, often operating on small, localized, and heterogeneous datasets that vary across users or environments. This limitation undermines both the adoption and adaptation of traditional CL methods in decentralized setups, as they are not inherently designed to operate under low-data regimes.

The shift towards a distributed learning setup offers the potential to learn representations locally on unlabeled data, helping to address privacy concerns and promoting fairness by eliminating the need for centralized data sharing. Distributed CL enables pervasive applications by facilitating on-device learning \cite{dong2021federated,seo2024relaxed, tan2022federated}, even in IoT battery-powered systems \cite{pham2022energy}.
Nevertheless, a critical gap remains in assessing the computational demands associated with training CL models at the edge/fog—including energy consumption, data requirements, and memory footprint—for deployment on devices with limited processing capabilities, even beyond the scope of distributed learning. Several real-world deployments illustrate these challenges. For instance, federated learning has been implemented on mobile devices to enhance next-word prediction without transmitting user data to centralized servers \cite{google_gboard_blog}. 
Beyond that, 
lightweight on-device models have been deployed to personalize notification rankings \cite{meta_notifications_blog}, while integrated learning capabilities have been introduced into processors to enable efficient AI workloads in mobile and IoT systems \cite{qualcomm_snapdragon_blog}. These examples underscore the need for a comprehensive investigation into the practical feasibility and applicability of CL methods under resource constraints.

\begin{figure*}[t]
    \centering
     \begin{subfigure}[b]{0.22\textwidth} 
\includegraphics[width=1\linewidth]{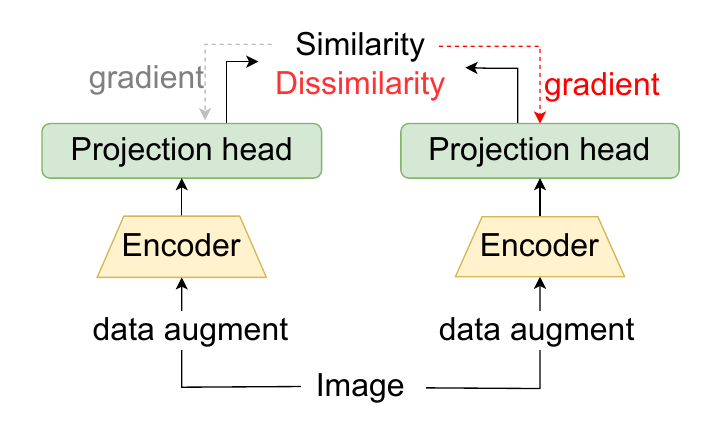}
\caption{SimCLR\label{fig:SimCLR}}
    \end{subfigure}
    \begin{subfigure}[b]{0.22\textwidth}
\includegraphics[width=1\linewidth]{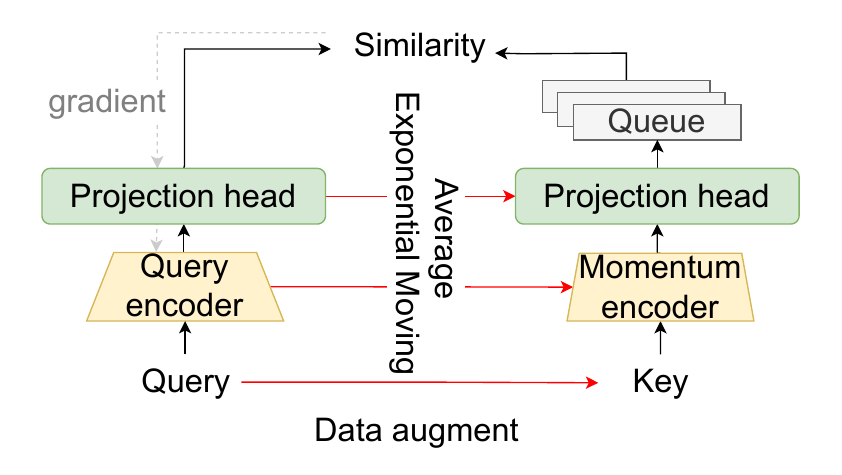}
\caption{MoCo\label{fig:MoCo}}
    \end{subfigure}
    \begin{subfigure}[b]{0.22\textwidth}
\includegraphics[width=1\linewidth]{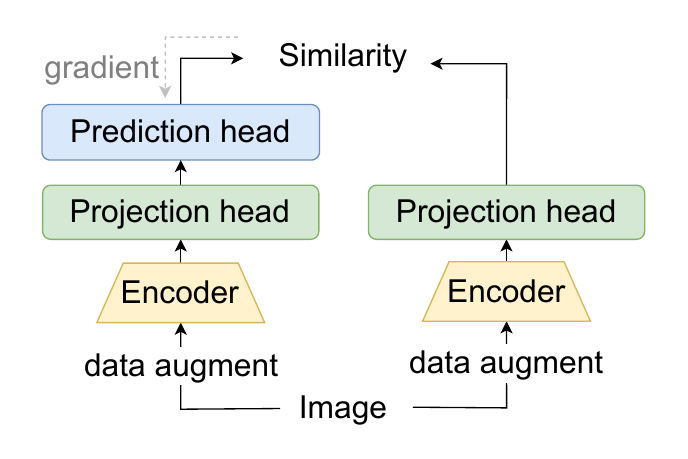}
    \caption{SimSiam\label{fig:simsiam}}
    \end{subfigure}
    \begin{subfigure}[b]{0.22\textwidth}
\includegraphics[width=1\linewidth]{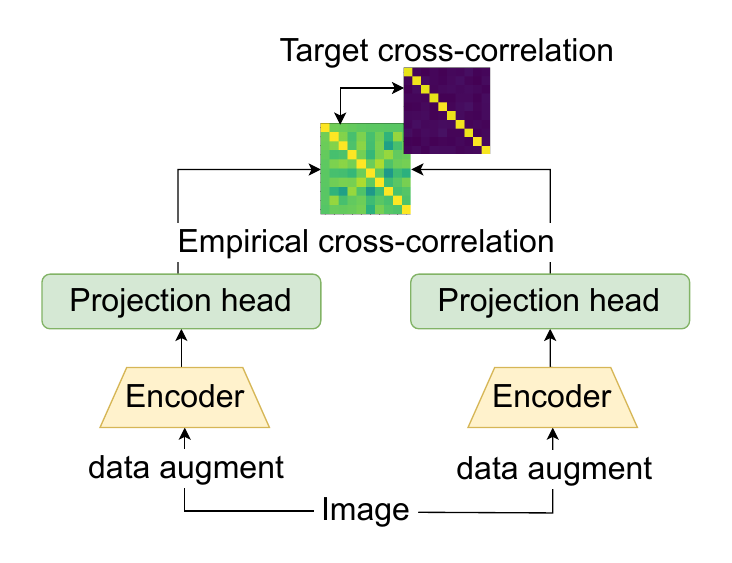}
\caption{Barlow Twins\label{fig:barlow}}    \end{subfigure}
    %\vspace{0.4cm}
    \caption{Illustration of the key operating principles for key representative CL frameworks. 
    }
    %\vspace{-0.3cm}
    \label{fig:CL_framework}  
 
\end{figure*}

To bridge this gap, this work conducts a systematic evaluation of prominent CL frameworks,
to measure their suitability for learning effective representations in resource-constrained environments. 
A key contribution of this study is the in-depth evaluation of energy consumption, a critical but often overlooked factor in the deployment of CL on edge/fog devices. Beyond energy profiling, we examine other essential resource requirements, including data availability and memory utilization, to gain insights into these frameworks' resulting trade-offs and computational demands.
Our results indicate that SimCLR emerges as the most energy-efficient CL framework, exhibiting the lowest energy usage across different data regimes.

While data availability represents a central limitation in such scenarios, it is not the only constraint; many edge/fog devices also lack the computational capacity to support large NN architectures for CL. To address this, we extend our analysis to include lightweight models that are better suited for deployment in environments with stringent computational and memory limitations.
Notably, despite having the largest parameter count among the models evaluated, ResNet-18 demonstrates the most favorable energy-to-accuracy ratio compared to its counterparts.
Overall, our findings aim to provide practical guidelines for designing more efficient and resource-aware CL solutions, thereby expanding their applicability to a broader range of edge/fog computing platforms, including low-power terminals.

\section{Background and Related Work}
\label{sec:background_related}

\subsection{Self-Supervised Contrastive Learning}

Various pretext tasks have been proposed to learn an optimal feature embedding,
including instance discrimination \cite{chen2020simple,he2020momentum}, self-distillation \cite{chen2021exploring, grill2020bootstrap}, and feature decorrelation \cite{zbontar2021barlow}. SimCLR and MoCo are examples of instance discrimination, which seeks to maximize the alignment between augmented views of the same sample while minimizing the similarity between augmented views of different images. These augmented views can be obtained by applying data augmentation operators such as cropping, rotation, or Gaussian blur to an original image. 

In SimCLR (Fig.~\ref{fig:SimCLR}), data augmentation is applied to each image sample to obtain two correlated views, which form a positive pair.
A neural network, comprising an encoder to extract representation vectors from the views and a small projection head to map these representations to the latent space, is then trained. A contrastive loss is used 
to maximize the correspondence between positive pairs and minimize it between negative pairs, i.e., the views generated from other images within the same batch.
SimCLR benefits from large batch sizes to provide a sufficient number of negative samples for improved performance. 

MoCo (Fig.~\ref{fig:MoCo}), in contrast, decouples the batch size from the number of negative pairs 
by employing a dual-network architecture. 
MoCo's limitation is no longer the batch size, but the need for a large queue to accumulate a high number of negative pairs and store both networks in memory. SimCLR and MoCo utilize the InfoNCE loss function \cite{oord2019} and typically a $128$-dimensional feature embedding to achieve effective representations. 

BYOL and SimSiam (Fig.~\ref{fig:simsiam}), both examples of self-distillation tasks, together with Barlow Twins, a feature correlation-based approach, employ different mechanisms to eliminate the dependency on negative samples.
 BYOL uses an 
 online network that incorporates a predictor, which enhances the target network by attempting to align its output with that of the target network.
 SimSiam also employs a predictor;  
however, rather than utilizing past interactions of the online network as a target, it relies on a single network that serves as both the online and target network, with no gradient computed with respect to the target network's weights. In this study, a $2048$-dimensional feature embedding is used, along with a prediction MLP hidden layer set to $1/4$ of the output dimension, ensuring a more stable and robust structure.

In Barlow Twins (Fig.~\ref{fig:barlow}), the loss function is based on a cross-correlation matrix computed between 
 batches of embeddings. 
 This cross-correlation loss is memory-intensive, as it requires storing two batches of embeddings to compute the matrix. Unlike approaches that utilize the InfoNCE loss function, this method benefits from high-dimensional feature embeddings, leading to improved performance.

\subsection{Contrastive Learning at the Edge}

Locally deploying traditional CL approaches is impractical due to the high computational cost caused by deep and complex models, large batch sizes (or extensive memory queues), and prolonged training times to achieve optimal performance. To mitigate these expensive training requirements, a common strategy involves leveraging pre-trained models to transfer knowledge to other models through knowledge distillation (KD)
\citep{gao2022disco}. Although this method can be effective in certain cases, its applicability to real-world scenarios and repeated experimentation involving resource-constrained devices remains questionable. 

Various studies have explored strategies to adapt CL frameworks for edge devices, by reducing the dataset size \cite{Kinakh_Taran_Voloshynovskiy_2021} or employing NNs with lower computational capacity \cite{Lin_Ding_Cao_Zheng_2023, Shi_Zhang_Tang_Zhu_Li_Guo_Zhuang_2022}. Memory-efficient sampling strategies \cite{joshi2024data} and computationally efficient augmentation techniques \cite{wen2021toward} can optimize model training, thereby reduce resource consumption on target devices. In this latter,
their findings indicate that employing natural data augmentations during training effectively decouples spurious correlations between the representations of positive samples, leading to more efficient feature learning.

ScatSimCLR architecture \cite{Kinakh_Taran_Voloshynovskiy_2021} aims to address the challenges of small-scale datasets in SSL by replacing the conventional ResNet with a customized embedding layer. Careful hyperparameter tuning is shown to improve CL performance when lightweight models are trained \cite{Shi_Zhang_Tang_Zhu_Li_Guo_Zhuang_2022}—an approach also followed in \cite{Lin_Ding_Cao_Zheng_2023}.
The work in \cite{qian2022makes} focuses on wearable-based applications, which inherently operate under small-scale and resource-constrained conditions. Their study explores both the algorithmic and task-specific strategies to optimize CL for human activity recognition.

Despite the considerable advancements in the CL field,
we identify a notable gap in research addressing the energy consumption aspects of SSL.
This endeavor is of extreme importance to understand the bounds and limitations of CL methods directly implemented on low-powered and resource-constrained edge/fog devices and to guide future research efforts toward possible optimization methods. 
Thus, our study differs significantly from previous works by investigating the impact of training various CL frameworks on energy consumption—an aspect that has been largely overlooked or insufficiently explored in prior research. 
Furthermore, we assess the capabilities of different CL strategies to extract rich representations under limited training data settings, and elaborate not only on their performance implications but also on the corresponding energy costs, enabling a comprehensive analysis of the trade-off between accuracy gains and energy expenditure.
To facilitate an in-depth evaluation, we consider a variety of NN architectures specifically designed or adapted for resource-constrained environments, such as edge devices and low-power IoT hardware. These architectures were selected based on their reduced parameter count and their ability to maintain competitive performance despite hardware limitations.

\section{Methodology}
\label{sec:learning_edge}

We focus on two key measures for evaluating the practical application of CL at edge/fog environments: \textit{i}) energy consumption, a critical aspect for resource-constrained devices, and \textit{ii}) quality of representations, since the success of learning relies on the generalization of the representation.

\subsection{Measuring Energy Consumption}
\label{sec:learning_edge:energy_consumption}

Defining optimal results solely based on accuracy or similar performance metrics without considering computational efficiency is not sustainable in the longer term \cite{schwartz2020green}. 
One approach to addressing this issue is the integration
of an energy measuring mechanism,
which enables the identification of an optimal trade-off between performance and energy consumption. 
Energy efficiency is commonly assessed using floating-point operations (FLOPs). However, prior research has demonstrated that neither the number of trainable parameters nor the FLOP count exhibits a linear relationship with energy consumption \cite{tripp2024measuring}. This finding is reinforced by \cite{asperti2021dissecting}, which highlights a significant discrepancy between the efficiency inferred from the FLOP counts and the actual performance observed on modern hardware accelerators.
An alternative approach involves the use of direct 
energy measurement tools to obtain more realistic energy consumption estimates.
In this study, we focus on measuring computational energy consumption, as it represents the most significant component of overall energy usage. This has been demonstrated in earlier research~\cite{elia-ojcoms}.
To facilitate energy measurement, we utilize the
CodeCarbon\footnote{CodeCarbon available at: \url{https://github.com/mlco2/codecarbon}} package and estimate the total energy consumption as
\begin{eqnarray}\label{eq:energy}
    E = \frac{1}{N}\sum_{t_i=1}^{N}P_{\text{CPU}}(t_i) + P_{\text{GPU}}(t_i) + P_{\text{memory}}(t_i), 
\end{eqnarray}
where $P_{\text{CPU}}, P_{\text{GPU}}$, and $P_{\text{memory}}$ denote the hardware power consumption of the CPU, GPU, and RAM, respectively, as a function of the measurement period\footnote{The measurement period depends on the used monitoring tool (e.g., CodeCarbon) and the hardware employed.} $t_i = 1, \dots, N$.

\subsection{Measuring Quality of Representation}
\label{sec:learning_edge:quality_representation}

The acquisition of general-purpose representations from CL frameworks has led to several works aimed at measuring the quality of learned representations \cite{huang2021towards, wang2020understanding}. 
In this study, we focus on two such metrics, namely \emph{alignment} and \emph{variability collapse index (VCI)}.
It is worth noting that during our experiments a wider set of metrics (e.g., uniformity, divergence, and intra-class distance) have been assessed but did not provide any significant insights.
The selected metrics enable a quantitative assessment of whether different CL frameworks and NN architectures produce meaningful representations.
Specifically, alignment serves as a pretext task-related metric that can be adopted without any kind of annotation. On the other hand, VCI is a downstream-task-related metric that requires the use of labels. Combined these two metrics provide valuable insights on the overall distribution of representations in the latent space. 
A detailed description of each metric is provided below.

\vspace{0.05cm}

\noindent \textbf{Alignment} quantifies the similarity between positive augmentations in CL frameworks. 
Formally, it is defined as the average squared distance between the augmented views
\begin{eqnarray}\label{eq:align}
    \mathcal{L}_{align}=\frac{1}{\left| \mathcal{D} \right|}\sum_{x_j \in X}^{}{\left\| f\left( A_1\left( x_j \right)\right) -f\left( A_2\left( x_j \right) \right)  \right\|^2},
\end{eqnarray}
where $f(\cdot)$ denotes the feature extractor, and $A_1(x_j)$ and $A_2(x_j)$ represent two augmented versions of the sample $x_j$ belonging to the training set $\mathcal{D}$. 
This distance is expected to be intrinsically minimized during training, ensuring that similar samples are mapped to nearby points in the latent space.

\vspace{0.05cm}

\noindent \textbf{Variability Collapse Index (VCI)} aims to assess the transferability of NNs, i.e., whether a pre-trained model is capable of performing well on different downstream tasks \cite{seo2024relaxed, xu2023quantifying}. It is motivated by the concept of \textit{neural collapse}, 
%\ff{\cite{papyan2020prevalence_neural_collapse}},
which suggests that, as training progresses, the latent representation of each sample converges to its corresponding class mean. While neural collapse can be beneficial for the pretext task, it may also negatively affect the model's transferability to new downstream tasks. In the context of CL, excessively collapsed representations reduce variability, potentially hindering the model's generalization capability.  

Let  $h_{j}^{(k)} = f( x_{j}^{(k)})$ denote the feature vector of the $j$th data sample $x_{j}^{(k)} \in \mathcal{D}_k$  belonging to the $k$th class, where $k = 1, \dots, K$ with $K$ being the total number of classes. The within-class ($\Sigma_{W}$)  and between-class ($\Sigma_{B}$)  covariance matrices capture the variability of representations within the same and across different classes, respectively 
\cite{xu2023quantifying}. 
Using these second-order statistics together with the overall covariance matrix $\Sigma_{T} = \Sigma_{B} + \Sigma_{W}$, VCI can be defined as 
\begin{eqnarray}
    \mathrm{VCI} = 1 - \frac{\text{Tr}\left[ \Sigma_{T}^{\dagger} \Sigma_{B}\right]}{\text{rank}\left( \Sigma_{B}^{} \right)}.
\end{eqnarray}
Note that this metric is invariant to invertible
linear transformations of the latent space, making it a useful measure of transferability
to downstream tasks \cite{xu2023quantifying}. In this study, we will rely on both alignment and VCI metrics to evaluate the quality of representations learned by various CL frameworks and NN architectures.

\section{Numerical Evaluation}
\label{sec:experiments}

Our evaluation is conducted across four state-of-the-art CL frameworks, namely SimCLR, MoCo, Barlow Twins, and SimSiam, using six different NN architectures, including ResNet-18, EfficientNet (B0–B2), SqueezeNet, 
and MobileNet.
These architectures were selected due to their 
relatively low parameter count and their
suitability for deployment on resource-constrained devices.

To enhance readability, we first present general results using ResNet-18 as a representative architecture. We then extend the analysis to additional architectures,
 offering comparative insights into their performance across various CL frameworks.

\subsection{Experimental Setup}
\label{sec:experiments:setup}

We employ a k-nearest neighbors (kNN) classifier to measure accuracy for $800$ epochs and use the following data transformations for generating augmented views, unless stated otherwise: (a) random cropping, (b) random Gaussian blur, (c) color dropping, (d) color distortion, and (e) random horizontal flipping.  
We run our experiments on the CIFAR-10 dataset with a batch size of $512$. For the experiments exclusively using ResNet-18, we adopt the model implementation provided by the \textit{Lightly} library.
When comparing ResNet-18 with other NN architectures, we use model implementations provided by the \textit{Torchvision} library to ensure consistency across architectures. 
To ensure fair comparison among different NN architectures and CL frameworks, we include a 2D convolutional layer to set the model output to a $512$-dimensional size, which is then projected according to  
the framework, as done in prior benchmark on CL \cite{huang2021towards}.  

The computational energy during training is assessed with the aid of CodeCarbon
library following \eqref{eq:energy}.\footnote{For all experiments, we use an Intel Xeon Silver 4314 CPU in constant mode for CPU energy estimation. A constant thermal design power of $135$W is assigned, so that the final allocated power is $67.5W$. For memory consumption, $\nicefrac{3}{8}$W per GB of available RAM is considered, which corresponds to $47$W.

Preliminary experiments were also conducted on devices with more constrained memory and GPU resources, showing trends consistent with those reported. However, results are not included due to the additional hyperparameter tuning required to fit models within constrained memory, thereby making the experimentation setup much larger.
}

For monitoring $P_{\text{GPU}}$, 
we employ the \textit{pynvml} library,
 which is integrated into the library. 
For measuring 
$P_{\text{CPU}}$ on Linux operating systems with Intel processors, we rely on the Intel Running Average Power Limit (RAPL).
Hardware monitoring is conducted at frequent time intervals, with a default sampling rate of $15$s. For $P_{\text{memory}}$, CodeCarbon assumes that every $8$GB of RAM consumes approximately $3$W of energy.
We captured the energy consumption for $20$ epochs, discarding the first $5$ epochs to control instability in the process initialization. During our experiments, we observed that the energy consumed per epoch remained relatively stable after this warm-up stage.
The final energy consumption is computed as the average of the remaining epochs.

\subsection{Data-Performance Trade-off}
\label{sec:experiments:data_performance}

We conduct a comprehensive series of experiments 
to evaluate the performance of the CL frameworks mentioned above in terms of classification accuracy and energy consumption across varying training data sizes. 
To emulate different data availability scenarios,
we randomly sample subsets comprising $2\%$, $5\%$, $10\%$, $20\%$, $40\%$, $60\%$, $80\%$, and $100\%$ of the entire dataset. We consider that every subset follows the same distribution as the original dataset, i.e., the ratio of samples per class is kept the same.
Scenarios using more than $40\%$ of the dataset represent training conditions typical of centralized, high-capacity servers, while those involving less than $20\%$ of the data mimic settings with limited data availability, as often encountered on 
distributed scenarios at the edge.

\begin{figure}[t!]
    \centering
    % This file was created with tikzplotlib v0.10.1.

\tikzset{
    type1/.style={
        execute at end picture={
            \coordinate (lower right) at (current bounding box.south east);
            \coordinate (upper left) at (current bounding box.north west);
        }
    },
    type/.style={
        execute at end picture={
            \pgfresetboundingbox
            \path (upper left) rectangle (lower right);
        }
    }
}

\begin{tikzpicture}[type1]
\definecolor{darkgray176}{RGB}{176,176,176}
\definecolor{green}{RGB}{0,128,0}
\definecolor{lightgray204}{RGB}{204,204,204}
\definecolor{violet}{RGB}{238,130,238}

\begin{axis}[
width=0.75\linewidth,
height=0.65\linewidth,
legend cell align={left},
legend style={
  fill opacity=0.8,
  draw opacity=1,
  text opacity=1,
  legend columns=4,
  at={(0.5,1.3)},
  anchor=south,
  draw=lightgray204,
  /tikz/every even column/.append style={column sep=0.3cm},
},
tick align=outside,
tick pos=left,
x dir=reverse,
x grid style={darkgray176},
xlabel={Proportion of dataset},
xmin=0, xmax=100,
xtick style={color=black},
y grid style={darkgray176},
ylabel={Energy per epoch (Wh)},
ymin=0, ymax=79.59,
ytick style={color=black}
]
\addplot [semithick, violet, dash pattern=on 2.55pt off 2.4pt, mark=*, mark size=2, mark options={solid}]
table {%
100 67.7
80 51
60 35.2
40 22.1
20 11.2
10 8.2
5 6.4
2 5.1
};

\addplot [semithick, blue, dash pattern=on 1.55pt off 2.4pt, mark=*, mark size=2, mark options={solid}]
table {%
100 75.8
80 57
60 39.5
40 24.7
20 13
10 9
5 6.7
2 5.2
};

\addplot [semithick, red, dash pattern=on 2.55pt off 2.4pt, mark=*, mark size=2, mark options={solid}]
table {%
100 61.6
80 46.4
60 31.7
40 19.8
20 10.7
10 7.8
5 6.1
2 5.1
};

\addplot [semithick, green, dash pattern=on 2.55pt off 2.4pt, mark=*, mark size=2, mark options={solid},]
table {%
100 73.5
80 55.5
60 38.6
40 24.2
20 13.2
10 10.7
5 7.4
2 5.6
};

\end{axis}

\begin{axis}[
width=0.75\linewidth,
height=0.65\linewidth,
axis y line=right,
legend columns=3,
legend style={
  fill opacity=0.8,
  draw opacity=1,
  text opacity=1,
  at={(1.2,1.25)},
  %anchor=north east,
  /tikz/every even column/.append style={column sep=0.3cm},
  draw=lightgray204
},
tick align=outside,
x dir=reverse,
x grid style={darkgray176},
xmajorgrids,
xmin=0, xmax=100,
xtick pos=left,
xtick style={color=black},
y grid style={darkgray176},
ylabel={Accuracy},
ymajorgrids,
ymin=20, ymax=92.9765,
ytick pos=right,
ytick style={color=black},
yticklabel style={anchor=west}
]
\addplot [semithick, violet, mark=*, mark size=2, mark options={solid}]
table {%
100 89.02
80 88
60 86.02
40 83.02
20 76.27
10 71.61
5 61.04
2 44.32
};
\addlegendentry{SimCLR}
\addplot [semithick, blue, mark=*, mark size=2, mark options={solid}]
table {%
100 89.62
80 88.42
60 86.94
40 82.74
20 73.88
10 69.02
5 58.96
2 40.07
};
\addlegendentry{MoCo}
\addplot [semithick, red, mark=*, mark size=2, mark options={solid}]
table {%
100 88.34
80 86.72
60 77.9
40 62
20 49.42
10 50.32
5 37.54
2 24.12
};
\addlegendentry{SimSiam}
\addplot [semithick, green, mark=*, mark size=2, mark options={solid}]
table {%
100 86.34
80 84.67
60 82.14
40 79.01
20 60.74
10 55.72
5 40.09
2 22.49
};
\addlegendentry{Barlow Twins}
\addplot [semithick, black]
table {%
0 0
};
\addlegendentry{Accuracy}
\addplot [semithick, black, dash pattern=on 2.55pt off 2.4pt]
table {%
0 0
};
\addlegendentry{Energy}
\end{axis}
\end{tikzpicture}
    \vspace*{-0.7cm}
    \caption{Energy consumption (dashed lines, left y-axis) and accuracy (solid lines, right y-axis) as the amount of available training data is reduced.}
    \label{fig:resnet18_data_energy}
    \vspace*{-0.1cm}
\end{figure}
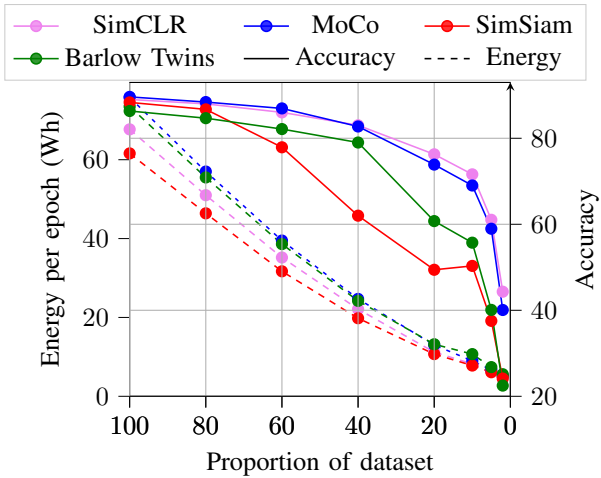

\begin{figure*}[t!]
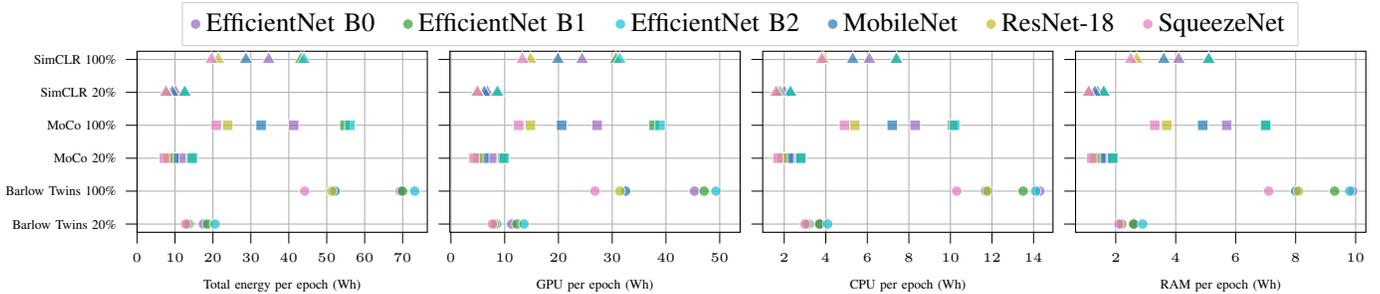

    \centering
    \include{img/tikz/total_energy}
    \vspace{-0.7cm}
     \caption{Measuring energy consumption components for SimCLR, MoCo, and Barlow Twins frameworks considering several NN architectures while the amount of data is reduced from $100\%$ to $20\%$.}
    %\vspace{-0.4cm}
\label{fig:resnet18_performance_energy_frameworks}
\end{figure*}

Fig. \ref{fig:resnet18_data_energy} displays the final test accuracy (solid lines) and energy consumption (dashed lines) for training a ResNet-18 model
under different data sizes ($x$-axis). 
While the decay in energy is almost linear with the number of training samples, the accuracy levels decrease similar to an inverse exponential curve for SimCLR and MoCo, and to a broken power-law pattern for SimSiam and Barlow Twins. This trend is observed across all evaluated CL frameworks, indicating a trade-off between energy consumption and test accuracy. 
Our results reveal that energy consumption is reduced by $\sim$ $83\%$ when training with only $20\%$ of the dataset, compared to full dataset utilization. However, this comes at the cost of accuracy degradation—approximately $21\%$ for SimCLR and MoCo, $43.64\%$ for Barlow Twins and $51.32\%$ for SimSiam, respectively.

Notably, Barlow Twins exhibits a significant performance drop when trained with less than $40\%$ of the data, while SimSiam undergoes a sharp decline in accuracy once the dataset size drops below $80\%$.
 Overall, 
 we notice that when going from $100\%$ to $2\%$ data regime, the drop in accuracy in Barlow Twins and SimSiam is almost $2\times$ higher than the drop in accuracy for SimCLR and MoCo. This remarkable sensitivity to dataset size suggests that Barlow Twins and SimSiam are considerably more vulnerable to data scarcity, with even modest reductions in training data resulting in a rapid degradation in model performance. 
 Specifically, 
 SimSiam fails to compensate for the absence of negative pairs when the amount of data is reduced, which leads to a collapse of the model—a finding that is consistent with the observations reported in \cite{li2022understanding}.

\subsection{Energy Profiling}
\label{sec:experiments:performance}
In the previous section, we analyzed how different loss functions behave in terms of energy consumption and accuracy under reduced training data conditions. 
This allowed us to investigate the adaptability of these frameworks to scenarios characterized by limited data availability, as typically encountered in resource-constrained environments. 
However, data availability is only one aspect of the challenge. Devices operating in such environments often lack the computational capabilities to support large models such as ResNet-18. To address this, we hereby extend our analysis to include lightweight NN architectures. 

\subsubsection{CL Frameworks}
We assess the energy consumption profiles of different CL frameworks when paired with lightweight models under conditions of limited data availability.
Specifically, we compare the energy levels consumed by the GPU, CPU, and RAM during training on both the full dataset and a reduced subset comprising $20\%$ of randomly selected samples.
Fig. \ref{fig:resnet18_performance_energy_frameworks} presents the energy consumed (per epoch) 
for training each CL framework using different models and under these two data sizes.
As expected, training on a reduced dataset leads to lower energy consumption across all three CL frameworks, irrespective of the model used. More interesting, however, is the rate of reduction. Specifically, for SimCLR and MoCo, training on 100\% of the data requires approximately $2.5\times-3.5\times$ more energy than training on just 20\% of the data, whereas for Barlow Twins, the rate increases to approximately $3.5\times-4\times$.

SimCLR seems to be the most energy-efficient framework, consistently exhibiting the lowest energy usage regardless of the data regime. 
In particular, it is more RAM efficient than MoCo and Barlow Twins as it does not require any additional memory structure, such as MoCo's momentum queue or Barlow Twins' large correlation matrices. While this efficiency becomes especially apparent when training with $100\%$ of the data,
it remains evident even with smaller data volumes.
SimCLR's CPU energy consumption follows a similar trend as its RAM usage. 
In terms of GPU energy, SimCLR and MoCo show comparable performance.
Conversely, the CPU and RAM consumption in MoCo scales with the dataset size, an indication that is directly related to its reliance on the momentum queue. At $20\%$ of the data, the queue size and extra computations are still present, but they have a smaller impact compared to the overall training workload.

Barlow Twins consumes higher energy for both data sizes ($20\%$ and $100\%$), making it the most computationally expensive among the evaluated frameworks. 
Analyzing the GPU values in Fig. \ref{fig:resnet18_performance_energy_frameworks}, the high GPU energy consumption of Barlow Twins compared to the other CL frameworks can mainly be explained by the difference in the projector dimensionality, as it is highly susceptible to the reduction of the representation dimension. This is compounded by the inclusion of a third MLP projection head and the computationally intensive loss function, which involves large matrix operations and gradient computations. 
Furthermore, storing this large correlation matrix necessitates significant RAM memory, resulting in elevated energy expenditure.

\subsubsection{Neural Network Architectures}
To further investigate the interplay between architecture complexity and energy usage, we analyze a selection of lightweight models commonly used in resource-limited environments. 
Contrary to the general intuition that a high number of parameters yields high energy consumption, the results from Fig. \ref{fig:resnet18_performance_energy_frameworks} and Table \ref{tab:architectures_energy_acc} suggest otherwise. 
ResNet-18, while having the largest parameter count among the tested models,
%ResNet-18 
exhibits the best accuracy and the most favorable energy-to-accuracy ratio.

In contrast, EfficientNet variants (B0–B2) exhibit the highest energy consumption, even when training with only $20\%$ of the dataset. This is especially pronounced when paired with MoCo and Barlow Twins frameworks.  
A detailed breakdown of energy usage reveals that EfficientNet consistently requires more GPU, CPU, and RAM energy than ResNet-18, despite having fewer parameters. This can be attributed to its practical inefficiency in training on GPU hardware, as previously reported in \cite{masters2021making}, which exacerbates its total energy footprint.
As for the energy consumption of MobileNet, it demonstrates lower energy consumption than EfficientNet, yet still exceeds that of  
ResNet-18. We suspect that this is due to the computational inefficiency of depthwise separable convolutions, which involve fragmented operations that are not well-optimized for GPU execution 
\cite{lee2019energy}.

SqueezeNet proved to be the most energy-efficient architecture across all training data sizes (albeit with a reduction in accuracy).
SqueezeNet uses fire modules, which are blocks composed of different convolution layers used to reduce 
the number of parameters while ensuring reasonable accuracy, making it well-suited for devices with limited resources. 
However, during training, SqueezeNet exhibited a tendency toward representation collapse, i.e., producing indistinct embeddings in the latent space. This suggests limited feature extraction capabilities, consistent with findings in \cite{yang2021comparative},
which directly affects representation quality.
Thus, although lightweight models may appear attractive for low-resource settings, our results indicate that architectural compactness does not necessarily translate to lower energy consumption or improved performance.

\begin{table}[t!]
 \renewcommand{\arraystretch}{1.1}
\caption[Caption for LOF]{Energy expenditure for training different models with CL frameworks (20\% and 100\% of training data).\footnotemark}
\resizebox{\columnwidth}{!}{%
\begin{tabular}{|llc|cc|cc|cc}
\cline{1-7} 
& & & \multicolumn{2}{c|}{\textbf{20\% Data}} & \multicolumn{2}{c|}{\textbf{100\% Data}} \\
 \textbf{CL Framework} & \textbf{Models} & \textbf{\# Params} & \textbf{Acc (\%)} & \textbf{Energy} & \textbf{Acc (\%)} & \textbf{Energy} \\
 & & & & \textbf{(per epoch $\pm$ std)} & & \textbf{(per epoch $\pm$ std)} \\
\hline 
\multirow{6}{*}{SimCLR} &
  ResNet-18 &
  11.5M &
  {\textbf{67.32}} &
  {\underline{7.8}} $\pm$
  0.28 &
  {\textbf{79.36}} &
  {\underline{21.4}} $\pm$
  0.83 \\ 
 &
  SqueezeNet &
  1.06M &
  {44.12} &
  {\textbf{7.6}} $\pm$
  0.27 &
  {57.75} &
  {\textbf{19.6}} $\pm$
  0.76 \\ 
 &
  EfficientNet B0 &
  4.99M &
  {\underline{61.74}} &
  {10.2} $\pm$
  0.36 &
  {\underline{66.23}} &
  {34.7} $\pm$
  1.29 \\ 
 &
  EfficientNet B1 &
  7.5M &
  {59.25} &
  {12.5} $\pm$
  0.45 &
  {65.94} &
  {43.2} $\pm$
  1.59 \\ 
 &
  EfficientNet B2 &
  8.75M &
  {59.47} &
  {12.6} $\pm$
  0.45 &
  {66.19} &
  {44} $\pm$
  1.6 \\ 
 &
  MobileNet &
  3.79M &
  {59.58} &
  {9.3} $\pm$
  0.33 &
  {66.0} &
  {28.7} $\pm$
  1.08 \\ \hline
\cline{1-7} 
%
%
% % % % MOCO
%
%
 \multirow{6}{*}{MoCo} &
  ResNet-18 &
  11.5M &
  {\textbf{68.69}} &
  {\underline{8.2}} $\pm$
  0.3 &
  {\textbf{81.4}} &
  {\underline{23.9}} $\pm$
  0.93 \\ 
 &
  SqueezeNet &
  1.06M &
  {36.72} &
  {\textbf{7.2}} $\pm$
  0.26 &
  {\underline{66.33}} &
  {\textbf{20.9}} $\pm$
  0.84 \\ 
 &
  EfficientNet B0 &
  4.99M &
  {\underline{56.51}} &
  {11.7} $\pm$
  0.42 &
  {65.01} &
  {41.3} $\pm$
  1.5 \\ 
 &
  EfficientNet B1 &
  7.5M &
  {53.01} &
  {14.3} $\pm$
  0.51 &
  {64.16} &
  {54.8} $\pm$
  1.97 \\ 
 &
  EfficientNet B2 &
  8.75M &
  {55.61} &
  {14.6} $\pm$
  0.53 &
  {64.88} &
  {56.1} $\pm$
  2.04 \\ 
 &
  MobileNet &
  3.79M &
  {54.56} &
  {9.7} $\pm$
  0.35 &
  {62.43} &
  {32.7} $\pm$
  1.23 \\ \hline
\cline{1-7} 
%
%
% % % % BT
%
%
\multirow{6}{*}{Barlow Twins} &
  ResNet-18 &
  22.7M &
  {\textbf{61.88}} &
  {\underline{13.4}} $\pm$
  0.47 &
  {\textbf{76.04}} &
  {\underline{51.3}} $\pm$
  1.79 \\  
 &
  SqueezeNet &
  12.29M &
  {10.0} &
  {\textbf{12.8}} $\pm$
  0.45 &
  {10.0} &
  {\textbf{44.2}} $\pm$
  1.6 \\  
 &
  EfficientNet B0 &
  16.22M &
  {\underline{35.81}} &
  {17.5} $\pm$
  0.63 &
  {58.71} &
  {69.4} $\pm$
  2.45 \\  
 &
  EfficientNet B1 &
  18.73M &
  {33.02} &
  {18.6} $\pm$
  0.66 &
  {57.86} &
  {70} $\pm$
  2.5 \\  
 &
  EfficientNet B2 &
  19.98M &
  {33.67} &
  {20.6} $\pm$
  0.72 &
  {\underline{69.23}} &
  {73.2} $\pm$
  2.62 \\  
 &
  MobileNet &
  15.02M &
  {31.11} &
  {13.8} $\pm$
  0.49 &
  {65.09} &
  {52.2} $\pm$
  1.86 \\ \hline
\cline{1-7} 
\end{tabular}
\label{tab:architectures_energy_acc}
}
%\vspace*{-0.3cm}
\end{table}
\footnotetext{The difference in the number of parameters of $11.6$M in Barlow Twins compared to the other two frameworks is mainly attributed to the addition of a third MLP projection head and the use of a high-dimensional representation of $2048$ (as opposed to $128$ in the others).}

\subsection{Quality of Representations} 
\label{sec:performance representation}
To obtain general-purpose representations, it is essential to employ evaluation mechanisms that go beyond traditional accuracy assessment. 
 
As introduced in Section \ref{sec:learning_edge:quality_representation}, the alignment captures the similarity between the representations of the augmented views of the same input sample. Thus, low alignment scores indicate that these representations lie closer in the feature space. 
In contrast, the VCI score 
quantifies the degree of representational collapse, providing a robust indicator of representation transferability and diversity. 
 In both cases, lower values indicate better-quality representations.
As shown in Table \ref{tab:alignment_VCI},
both alignment and VCI scores improve as the quantity of training data increases, a trend observed across all evaluated models. Among these, ResNet-18 consistently achieves the lowest alignment and VCI scores, indicating high-quality and stable representations regardless of the training data size. This further corroborates the findings from previous experiments, where ResNet-18 demonstrated superior performance in both accuracy and energy-efficiency trade-offs.
We hypothesize that ResNet-18’s superior performance may be attributed, in part, to its compatibility with the hyperparameters used in training—such as learning rate, weight decay, and momentum—as originally tuned for CL frameworks. While limited hyperparameter tuning was performed for the other models, no significant performance improvements were observed.
SqueezeNet, on the other hand, 
exhibits the highest alignment and VCI values in all experiments, suggesting a susceptibility to representation collapse.

Additionally, our analysis of EfficientNet (B1-B2), and MobileNet models trained on a reduced dataset ($20\%$) reveals that improvements in accuracy do not necessarily correlate with enhancements in other representation quality metrics. Specifically, despite a slight increase in accuracy from EfficientNet B1 to EfficientNet B2 and MobileNet, the alignment score marginally deteriorates while the VCI score improves. This observation highlights the potential pitfalls of relying solely on a single metric to assess representation quality and underscores the need for multi-metric evaluation approaches in representation learning. An in-depth analysis of potential causal dependencies between the alignment and VCI metrics is beyond the scope of this study and will be addressed in future work.

\begin{table}[t!]
\renewcommand{\arraystretch}{1.1}
\centering
\footnotesize
\caption{Representation quality assessment for models trained with the SimCLR framework.}
\resizebox{0.85\columnwidth}{!}{%
\begin{tabular}{|cc|c|c|c|}
\hline
\textbf{Model}             & \textbf{\% Data} & \textbf{Accuracy} & \textbf{Alignment} & \textbf{VCI} \\ \hline
\multirow{2}{*}{ResNet-18} & 100\%            & \textbf{79.36}             & \textbf{0.3271}             & \textbf{0.4814}       \\ %\cline{2-5} 
 &   20\%  & \underline{67.32} & \underline{0.3597} & \underline{0.5431} \\ 
 \cline{1-5} 
\multirow{2}{*}{SqueezeNet}   & 100\% & 57.75  & 0.5188 & 0.729  \\ %\cline{3-6} 
 & 20\%  & 44.12 & 0.5738 & 0.7812 \\ 
 \cline{1-5}
 \multirow{2}{*}{EfficientNet B0} & 100\% & 66.23 & 0.4116 & 0.6236 \\% \cline{3-6} 
 &  20\%  & 61.74 & 0.4382 & 0.6407 \\ 
 \cline{1-5}
 \multirow{2}{*}{EfficientNet B1} & 100\% & 65.94 & 0.4027 & 0.6061 \\ %\cline{3-6} 
 &   20\%  & 59.25 & 0.4351 & 0.6913 \\ 
 \cline{1-5} 
 \multirow{2}{*}{EfficientNet B2} & 100\% & 66.19 & 0.4002 & 0.6063 \\ %\cline{3-6} 
 &  20\%  & 59.47 & 0.4392 & 0.6551 \\
 \cline{1-5} 
 \multirow{2}{*}{MobileNet}    & 100\% & 66.01 & 0.4156 & 0.6278 \\% \cline{3-6} 
 &  20\%  & 59.58 & 0.4453 & 0.6476 \\ \hline
\end{tabular}%
}
\label{tab:alignment_VCI}
%\vspace*{-0.3cm}
\end{table}

\subsection{Impact of Transformations}
\label{sec:experiments:transformations}

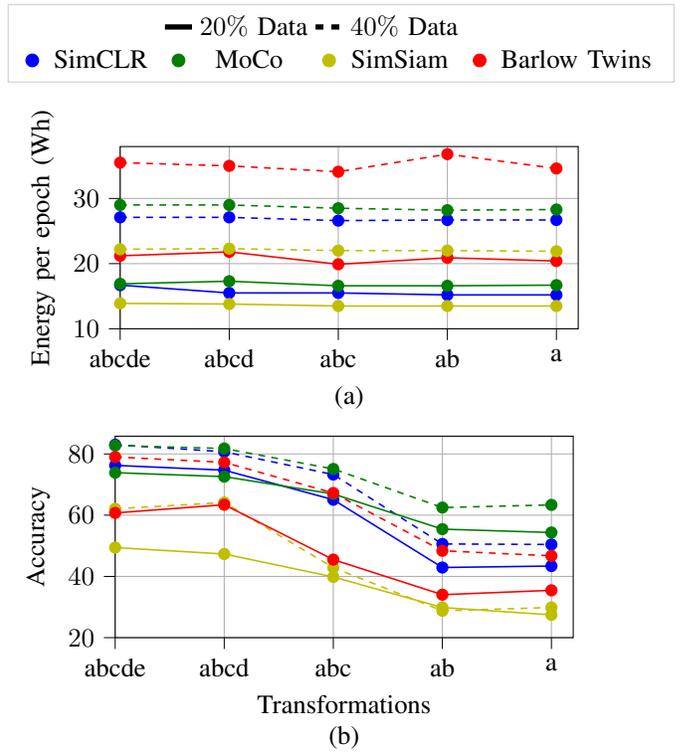
\begin{figure}[t!]
\centering
\hspace{-2.5em}
\begin{subfigure}[b]{0.45\textwidth}
    \definecolor{darkgray176}{RGB}{176,176,176}
\definecolor{goldenrod1911910}{RGB}{191,191,0}
\definecolor{green01270}{RGB}{0,127,0}
\definecolor{lightgray204}{RGB}{204,204,204}
\definecolor{green01270}{RGB}{0,127,0}

\begin{tikzpicture} 
\begin{axis}[
hide axis,
width=0.5\linewidth,
height=1in,%0.0\linewidth,
at={(0,0)},
xmin=10,
xmax=50,
ymin=0,
ymax=0.4,
legend style={fill opacity=1,
            draw opacity=1, 
            text opacity=1,
            draw=lightgray204,            
            legend columns=4,
    anchor=north,
    at={(0., 0.)},
legend image post style={scale=0.6}
    % font=\small,
    % legend image post style={scale=0.6}
}
]

% \addlegendimage{white, opacity=1, only marks}
\addlegendimage{white, dashed, line width=2pt}
\addlegendentry{}

\addlegendimage{black,  line width=2pt}
\addlegendentry{$20\%$ Data}

\addlegendimage{black, dashed, line width=2pt}
\addlegendentry{$40\%$ Data}

% \addlegendimage{white, opacity=1, only marks}
\addlegendimage{white, dashed, line width=2pt}
\addlegendentry{}%{$40\%$ Data}

% \addlegendimage{blue, line width=2pt};
% \addlegendentry{SimCLR \hspace{1em}};

% \addlegendimage{blue,  line width=2pt}
\addlegendimage{blue,mark=*, only marks, mark options={scale=2}};
\addlegendentry{SimCLR \hspace{1em}};

% \addlegendimage{green,  line width=2pt}
\addlegendimage{green01270,mark=*, only marks, mark options={scale=2}};
\addlegendentry{MoCo \hspace{1em}};

% \addlegendimage{goldenrod1911910,,  line width=2pt}
\addlegendimage{goldenrod1911910,mark=*, only marks, mark options={scale=2}};
\addlegendentry{SimSiam \hspace{1em}};

% \addlegendimage{red,  line width=2pt}
\addlegendimage{red,mark=*, only marks, mark options={scale=2}};
\addlegendentry{Barlow Twins \hspace{1em}};

\coordinate (legend) at (axis description cs:0.0,0.0);

\end{axis}

\end{tikzpicture}
    % \caption{}
    \vspace{-0.6cm}
\end{subfigure}
\begin{subfigure}[b]{0.47\textwidth}
    \tikzset{
    type1/.style={
        execute at end picture={
            \coordinate (lower right) at (current bounding box.south east);
            \coordinate (upper left) at (current bounding box.north west);
        }
        %/tikz/every even column/.append style={column sep=0.3cm}
        %/tikz/every even column/.style={column sep=2pt},
        %/tikz/every even row/.style={row sep=2pt}
    },
    type/.style={
        execute at end picture={
            \pgfresetboundingbox
            \path (upper left) rectangle (lower right);
        }
    }
}

\begin{tikzpicture}[type1]

\definecolor{darkgray176}{RGB}{176,176,176}
\definecolor{goldenrod1911910}{RGB}{191,191,0}
\definecolor{green01270}{RGB}{0,127,0}
\definecolor{lightgray204}{RGB}{204,204,204}

\begin{axis}[
width=0.9\linewidth,
height=0.47\linewidth,
legend cell align={left},
legend style={fill opacity=0.8,
    draw opacity=1, 
    text opacity=1, 
    at={(1.25,1.2)},    
    anchor=center,
    legend columns=4,
    /tikz/every even column/.append style={column sep=0.3cm},
    draw=lightgray204
    },
tick align=outside,
tick pos=left,
x grid style={darkgray176},
xlabel style={align=center},
xlabel={{(a)}},
xmajorgrids,
xmin=0, xmax=4.2,
xtick style={color=black},
xtick={0,1,2,3,4},
xtick={0,1,2,3,4},
xtick={0,1,2,3,4},
xtick={0,1,2,3,4},
xtick={0,1,2,3,4},
xtick={0,1,2,3,4},
xtick={0,1,2,3,4},
xtick={0,1,2,3,4},
xtick={0,1,2,3,4},
xticklabels={abcde,abcd,abc,ab,a},
xticklabels={abcde,abcd,abc,ab,a},
xticklabels={abcde,abcd,abc,ab,a},
xticklabels={abcde,abcd,abc,ab,a},
xticklabels={abcde,abcd,abc,ab,a},
xticklabels={abcde,abcd,abc,ab,a},
xticklabels={abcde,abcd,abc,ab,a},
xticklabels={abcde,abcd,abc,ab,a},
xticklabels={abcde,abcd,abc,ab,a},
y grid style={darkgray176},
ylabel={{Energy per  epoch (Wh)}},
ymajorgrids,
ymin=10, ymax=37.965,%1.2335
ytick style={color=black}
]
\addplot [semithick, blue, mark=*, mark size=2, mark options={solid}]
table {%
0 16.7
1 15.5
2 15.5
3 15.2
4 15.2
};
%\addlegendentry{SimCLR 20\%}
\addplot [semithick, green01270, mark=*, mark size=2, mark options={solid}]
table {%
0 16.9
1 17.3
2 16.6
3 16.6
4 16.7
};
%\addlegendentry{MoCo 20\%}
\addplot [semithick, goldenrod1911910, mark=*, mark size=2, mark options={solid}]
table {%
0 13.9
1 13.8
2 13.5
3 13.5
4 13.5
};
%\addlegendentry{SimSiam 20\%}
\addplot [semithick, red, mark=*, mark size=2, mark options={solid}]
table {%
0 21.2
1 21.8
2 19.9
3 20.9
4 20.4
};
%\addlegendentry{Barlow Twins 20\%}
\addplot [semithick, blue, dash pattern=on 2.55pt off 2.4pt, mark=*, mark size=2, mark options={solid}]
table {%
0 27.1
1 27.1
2 26.6
3 26.7
4 26.7
};
%\addlegendentry{SimCLR 40\%}
\addplot [semithick, green01270, dash pattern=on 2.55pt off 2.4pt, mark=*, mark size=2, mark options={solid}]
table {%
0 29
1 29
2 28.5
3 28.2
4 28.3
};
%\addlegendentry{MoCo 40\%}

\addplot [semithick, goldenrod1911910, dash pattern=on 2.55pt off 2.4pt, mark=*, mark size=2, mark options={solid}]
table {%
0 22.2
1 22.3
2 22
3 22
4 21.9
};
%\addlegendentry{SimSiam 40\%}
\addplot [semithick, red, dash pattern=on 2.55pt off 2.4pt, mark=*, mark size=2, mark options={solid}]
table {%
0 35.5
1 35
2 34.1
3 36.8
4 34.6
};
%\addlegendentry{Barlow Twins 40\%}
\end{axis}

\end{tikzpicture}
    % \caption{}
    \vspace{-0.6cm}
\end{subfigure}
\begin{subfigure}[b]{0.47\textwidth}
    \tikzset{
    type1/.style={
        execute at end picture={
            \coordinate (lower right) at (current bounding box.south east);
            \coordinate (upper left) at (current bounding box.north west);
        }
        %/tikz/every even column/.append style={column sep=0.3cm}
        %/tikz/every even column/.style={column sep=2pt},
        %/tikz/every even row/.style={row sep=2pt}
    },
    type/.style={
        execute at end picture={
            \pgfresetboundingbox
            \path (upper left) rectangle (lower right);
        }
    }
}

\begin{tikzpicture}[type1]

\definecolor{darkgray176}{RGB}{176,176,176}
\definecolor{goldenrod1911910}{RGB}{191,191,0}
\definecolor{green01270}{RGB}{0,127,0}
\definecolor{lightgray204}{RGB}{204,204,204}

\begin{axis}[
width=0.9\linewidth,
height=0.5\linewidth,
legend cell align={left},
tick align=outside,
tick pos=left,
x grid style={darkgray176},
xlabel style={align=center},
xlabel={{Transformations \\(b)}},
xmajorgrids,
xmin=0, xmax=4.2,
xtick style={color=black},
xtick={0,1,2,3,4},
xtick={0,1,2,3,4},
xtick={0,1,2,3,4},
xtick={0,1,2,3,4},
xtick={0,1,2,3,4},
xtick={0,1,2,3,4},
xtick={0,1,2,3,4},
xtick={0,1,2,3,4},
xtick={0,1,2,3,4},
xticklabels={abcde ,abcd ,abc ,ab ,a },
xticklabels={abcde ,abcd ,abc ,ab ,a },
xticklabels={abcde ,abcd ,abc ,ab ,a },
xticklabels={abcde ,abcd ,abc ,ab ,a },
xticklabels={abcde ,abcd ,abc ,ab ,a },
xticklabels={abcde ,abcd ,abc ,ab ,a },
xticklabels={abcde ,abcd ,abc ,ab ,a },
xticklabels={abcde ,abcd ,abc ,ab ,a },
xticklabels={abcde ,abcd ,abc ,ab ,a },
y grid style={darkgray176},
ylabel={Accuracy},
ymajorgrids,
ymin=20, ymax=85.7985,%24.6715
ytick style={color=black}
]
\addplot [semithick, blue, mark=*, mark size=2, mark options={solid}]
table {%
0 76.27
1 74.7
2 65.06
3 42.91
4 43.39
};
%\addlegendentry{SimCLR 20\%}
\addplot [semithick, blue, dash pattern=on 2.55pt off 2.4pt, mark=*, mark size=2, mark options={solid}]
table {%
0 83.02
1 80.71
2 73.28
3 50.6
4 50.43
};
%\addlegendentry{SimCLR 40\%}
\addplot [semithick, green01270, mark=*, mark size=2, mark options={solid}]
table {%
0 73.88
1 72.59
2 66.93
3 55.44
4 54.34
};
%\addlegendentry{MoCo 20\%}
\addplot [semithick, green01270, dash pattern=on 2.55pt off 2.4pt, mark=*, mark size=2, mark options={solid}]
table {%
0 82.74
1 81.78
2 75.1
3 62.49
4 63.36
};
%\addlegendentry{MoCo 40\%}
\addplot [semithick, goldenrod1911910, mark=*, mark size=2, mark options={solid}]
table {%
0 49.42
1 47.32
2 39.84
3 29.84
4 27.45
};
%\addlegendentry{SimSiam 20\%}
\addplot [semithick, goldenrod1911910, dash pattern=on 2.55pt off 2.4pt, mark=*, mark size=2, mark options={solid}]
table {%
0 62.09
1 64.14
2 42.88
3 28.77
4 29.85
};
%\addlegendentry{SimSiam 40\%}
\addplot [semithick, red, mark=*, mark size=2, mark options={solid}]
table {%
0 60.74
1 63.41
2 45.49
3 34.03
4 35.46
};
%\addlegendentry{Barlow Twins 20\%}
\addplot [semithick, red, dash pattern=on 2.55pt off 2.4pt, mark=*, mark size=2, mark options={solid}]
table {%
0 79.01
1 77.25
2 67.25
3 48.38
4 46.72
};
%\addlegendentry{Barlow Twins 40\%}
\end{axis}

\end{tikzpicture}
    % \caption{   }
\end{subfigure}
\vspace{-0.8cm}
\caption{Impact of transformations on (a) energy consumption and (b) accuracy performance.}
    %\vspace{-0.3cm}
    \label{fig:20_40_performance_energy}
\end{figure}
To assess the impact of data augmentation on performance and energy consumption, we incorporated a varying number of augmentation strategies into our training pipeline. This approach is motivated by the understanding that a key factor for the generalizability of CL in unsupervised settings is the \textit{concentration} of the learned representations \cite{huang2021towards}. Concentration can be effectively managed through the selection and quantity of augmentation strategies, often guided by domain-specific human expertise.
The specific augmentation strategies employed were mentioned in Section \ref{sec:experiments:setup}. 
For each image, two augmented views were created by sequentially applying these transformations.

As illustrated in Fig. \ref{fig:20_40_performance_energy}, 
increasing the number of transformations leads to only marginal variations in energy consumption, which remain negligible for most CL methods. Consequently, while the energy cost remains nearly constant, classification accuracy tends to improve as more augmentations are incorporated. This indicates that the inclusion of additional transformations serves as an \textit{energy-neutral} mechanism for enhancing model performance—particularly advantageous in resource-constrained scenarios where training data is limited.
Conversely, reducing the number of transformations reveals a noticeable sensitivity in certain CL frameworks. Notably, MoCo maintains relatively high accuracy levels in low-data regimes (i.e., $40\%$ and $20\%$ of the dataset), even when using only a single transformation strategy (“a”). In contrast, the accuracy gap between SimCLR and MoCo widens as the transformations are progressively removed, with MoCo demonstrating greater robustness under reduced augmentation settings.

SimCLR, Barlow Twins, and SimSiam exhibit significantly higher sensitivity to the reduction of transformations.
When only a single transformation is applied, accuracy drops below $51\%$ for all three frameworks when trained on $40\%$ of the dataset, and below $45\%$ under the $20\%$ data regime. SimSiam exhibits high vulnerability to augmentation scarcity, performing worse than Barlow Twins even when trained on a larger subset of the data. These findings are consistent with the results presented in Section \ref{sec:experiments:data_performance}, which highlighted SimSiam's reduced effectiveness in data-constrained scenarios. Overall, the insights derived from Fig. \ref{fig:20_40_performance_energy} emphasize the importance of carefully selecting augmentation strategies, especially in low-data or computationally constrained environments.

\section{Discussion}
This study addresses the deployment 
of CL frameworks in resource-constrained environments, particularly in scenarios where reliance on pre-trained models is impractical.
To enhance model performance under such limitations, we recommend the use of targeted data augmentation strategies, as well as CL frameworks that incorporate negative pairs in their foundation—since these have consistently demonstrated superior performance over their counterparts.
Strategies such as those presented in \cite{Shi_Zhang_Tang_Zhu_Li_Guo_Zhuang_2022}, can be explored to adequately guide the hyperparameter tuning 
of lightweight models. 
However, their effectiveness under limited training data remains uncertain, as our experiments indicate a tendency for overfitting when hyperparameter optimization is applied. This highlights the need for adaptive tuning methods suited for lightweight architectures in resource-limited settings, to gain insights into model suitability for various edge/fog scenarios.

Despite the promising outcomes of our study, several limitations should be acknowledged. Our energy consumption evaluation was conducted exclusively using the CodeCarbon library. While CodeCarbon offers accessible and reproducible estimates, it may not fully capture hardware-specific variations. Future research should therefore incorporate direct measurements on diverse edge/fog devices to assess both performance and energy expenditure in real-world conditions. This will enable a comprehensive evaluation of how hardware variability impacts training dynamics and the quality of learned representations.

Moreover, we aim to extend our study in order to evaluate the generalizability of learned representations across different domains and device types. In particular, exploring decentralized and collaborative training setups, such as self-supervised federated learning presents a promising direction. However, such scenarios introduce challenges like data heterogeneity (i.e., non-IID distributions), label imbalance, and client selection biases—all of which can significantly affect model convergence and fairness. Incorporating techniques like curriculum learning and informative data selection may help mitigate some of these challenges and improve learning outcomes in non-IID settings.

\section{Conclusions}
In this study, we investigate the feasibility of several prominent CL frameworks in resource-constrained environments. In particular, we address this gap by considering computational challenges that extend beyond model accuracy—namely, energy consumption, data scarcity, and memory footprint.
Our results emphasize that the energy cost associated with CL should not be overlooked, especially in real-world applications where battery life and computational budgets are critical factors. 
Additionally, we assess the capabilities of several lightweight models in learning useful representations with minimal computational burden. 
Among the frameworks evaluated, SimCLR demonstrated superior efficiency in energy consumption, while ResNet-18 emerged as the most balanced NN architecture in terms of energy-to-accuracy ratio.
%In doing so, we extend the unsustainable performance-oriented view, which usually only considers accuracy, to include the energy costs associated with training. We initiated our investigation by analyzing the effects on performance and energy consumption when training of four CL frameworks and showed that the decrease in performance is drastically disproportionate to the large decreases in energy expenditure during training of these frameworks when the amount of available data is reduced. %In our study, SimCLR appears to be the most energy-efficient framework that consumes the least energy regardless of the data regime. 
%Furthermore, we extend our investigation to models specifically designed or adapted for resource-constrained devices. In doing so, we not only analyze the accuracy and the different levels of CPU, GPU and RAM energy consumption, but we also evaluate the representation quality through alignment and VCI scores, so that we can gain further insights from the results obtained.
Our findings offer a foundation for future research into optimizing CL frameworks through informed design choices tailored to edge/fog computing. %\rmpp{environments}.

%%%%%%%%%%%%%%%%%%%%%%%%%%%%%%%%%%%%%%%%%%%%%%%%%%%%%%%%%%%%%%%%%%%%%%%%

%%% Use this command to include your bibliography file.

\bibliographystyle{IEEEtran}
\footnotesize
\bibliography{bibliography}

\begin{thebibliography}{10}
\providecommand{\url}[1]{#1}
\csname url@samestyle\endcsname
\providecommand{\newblock}{\relax}
\providecommand{\bibinfo}[2]{#2}
\providecommand{\BIBentrySTDinterwordspacing}{\spaceskip=0pt\relax}
\providecommand{\BIBentryALTinterwordstretchfactor}{4}
\providecommand{\BIBentryALTinterwordspacing}{\spaceskip=\fontdimen2\font plus
\BIBentryALTinterwordstretchfactor\fontdimen3\font minus \fontdimen4\font\relax}
\providecommand{\BIBforeignlanguage}[2]{{%
\expandafter\ifx\csname l@#1\endcsname\relax
\typeout{** WARNING: IEEEtran.bst: No hyphenation pattern has been}%
\typeout{** loaded for the language `#1'. Using the pattern for}%
\typeout{** the default language instead.}%
\else
\language=\csname l@#1\endcsname
\fi
#2}}
\providecommand{\BIBdecl}{\relax}
\BIBdecl

\bibitem{chen2020simple}
T.~Chen, S.~Kornblith, M.~Norouzi, and G.~Hinton, ``A simple framework for contrastive learning of visual representations,'' in \emph{International conference on machine learning}.\hskip 1em plus 0.5em minus 0.4em\relax PMLR, 2020, pp. 1597--1607.

\bibitem{he2020momentum}
K.~He, H.~Fan, Y.~Wu, S.~Xie, and R.~Girshick, ``Momentum contrast for unsupervised visual representation learning,'' in \emph{Proceedings of the IEEE/CVF conference on computer vision and pattern recognition}, 2020, pp. 9729--9738.

\bibitem{grill2020bootstrap}
J.-B. Grill, F.~Strub, F.~Altch{\'e}, C.~Tallec, P.~Richemond, E.~Buchatskaya, C.~Doersch, B.~Avila~Pires, Z.~Guo, M.~Gheshlaghi~Azar \emph{et~al.}, ``Bootstrap your own latent-a new approach to self-supervised learning,'' \emph{Advances in neural information processing systems}, vol.~33, pp. 21\,271--21\,284, 2020.

\bibitem{chen2021exploring}
X.~Chen and K.~He, ``Exploring simple siamese representation learning,'' in \emph{Proceedings of the IEEE/CVF conference on computer vision and pattern recognition}, 2021, pp. 15\,750--15\,758.

\bibitem{zbontar2021barlow}
J.~Zbontar, L.~Jing, I.~Misra, Y.~LeCun, and S.~Deny, ``Barlow twins: Self-supervised learning via redundancy reduction,'' in \emph{International conference on machine learning}.\hskip 1em plus 0.5em minus 0.4em\relax PMLR, 2021, pp. 12\,310--12\,320.

\bibitem{Shi_Zhang_Tang_Zhu_Li_Guo_Zhuang_2022}
H.~Shi, Y.~Zhang, S.~Tang, W.~Zhu, Y.~Li, Y.~Guo, and Y.~Zhuang, ``\BIBforeignlanguage{en}{On the efficacy of small self-supervised contrastive models without distillation signals},'' \emph{\BIBforeignlanguage{en}{Proceedings of the AAAI Conference on Artificial Intelligence}}, vol.~36, no.~2, p. 2225–2234, Jun. 2022.

\bibitem{gui_survey_2024}
\BIBentryALTinterwordspacing
J.~Gui, T.~Chen, J.~Zhang, Q.~Cao, Z.~Sun, H.~Luo, and D.~Tao, ``\BIBforeignlanguage{en}{A {Survey} on {Self}-supervised {Learning}: {Algorithms}, {Applications}, and {Future} {Trends}},'' Jul. 2024, arXiv:2301.05712 [cs]. [Online]. Available: \url{http://arxiv.org/abs/2301.05712}
\BIBentrySTDinterwordspacing

\bibitem{Lin_Ding_Cao_Zheng_2023}
\BIBentryALTinterwordspacing
W.~Lin, Y.~Ding, Z.~Cao, and H.-t. Zheng, ``\BIBforeignlanguage{en}{Establishing a stronger baseline for lightweight contrastive models},'' no. arXiv:2212.07158, Jul. 2023, arXiv:2212.07158 [cs]. [Online]. Available: \url{http://arxiv.org/abs/2212.07158}
\BIBentrySTDinterwordspacing

\bibitem{cao_improving_2024}
\BIBentryALTinterwordspacing
Y.-H. Cao and J.~Wu, ``\BIBforeignlanguage{en}{On {Improving} the {Algorithm}-, {Model}-, and {Data}- {Efficiency} of {Self}-{Supervised} {Learning}},'' Apr. 2024, arXiv:2404.19289 [cs]. [Online]. Available: \url{http://arxiv.org/abs/2404.19289}
\BIBentrySTDinterwordspacing

\bibitem{dong2021federated}
N.~Dong and I.~Voiculescu, ``Federated contrastive learning for decentralized unlabeled medical images,'' in \emph{International Conference on Medical Image Computing and Computer-Assisted Intervention}.\hskip 1em plus 0.5em minus 0.4em\relax Springer, 2021, pp. 378--387.

\bibitem{seo2024relaxed}
S.~Seo, J.~Kim, G.~Kim, and B.~Han, ``Relaxed contrastive learning for federated learning,'' in \emph{Proceedings of the IEEE/CVF Conference on Computer Vision and Pattern Recognition}, 2024, pp. 12\,279--12\,288.

\bibitem{tan2022federated}
Y.~Tan, G.~Long, J.~Ma, L.~Liu, T.~Zhou, and J.~Jiang, ``Federated learning from pre-trained models: A contrastive learning approach,'' \emph{Advances in neural information processing systems}, vol.~35, pp. 19\,332--19\,344, 2022.

\bibitem{pham2022energy}
Q.-V. Pham, M.~Le, T.~Huynh-The, Z.~Han, and W.-J. Hwang, ``Energy-efficient federated learning over uav-enabled wireless powered communications,'' \emph{IEEE Transactions on Vehicular Technology}, vol.~71, no.~5, pp. 4977--4990, 2022.

\bibitem{google_gboard_blog}
{Google AI}, ``Federated learning: Collaborative machine learning without centralized training data,'' \url{https://ai.googleblog.com/2017/04/federated-learning-collaborative.html}, 2017, blog post.

\bibitem{meta_notifications_blog}
{Meta Engineering}, ``Scaling personalization with on-device machine learning,'' \url{https://engineering.fb.com/2021/03/17/ml-applications/scaling-personalization-with-on-device-machine-learning/}, 2021, engineering blog.

\bibitem{qualcomm_snapdragon_blog}
{Qualcomm Technologies, Inc.}, ``Enabling on-device ai with snapdragon,'' \url{https://www.qualcomm.com/news/onq/2020/09/30/enabling-device-ai-snapdragon}, 2020, official Qualcomm blog.

\bibitem{oord2019}
\BIBentryALTinterwordspacing
A.~van~den Oord, Y.~Li, and O.~Vinyals, ``{Representation Learning with Contrastive Predictive Coding},'' 2019. [Online]. Available: \url{https://arxiv.org/abs/1807.03748}
\BIBentrySTDinterwordspacing

\bibitem{gao2022disco}
Y.~Gao, J.-X. Zhuang, S.~Lin, H.~Cheng, X.~Sun, K.~Li, and C.~Shen, ``Disco: Remedying self-supervised learning on lightweight models with distilled contrastive learning,'' in \emph{European Conference on Computer Vision}.\hskip 1em plus 0.5em minus 0.4em\relax Springer, 2022, pp. 237--253.

\bibitem{Kinakh_Taran_Voloshynovskiy_2021}
\BIBentryALTinterwordspacing
V.~Kinakh, O.~Taran, and S.~Voloshynovskiy, ``\BIBforeignlanguage{en}{Scatsimclr: self-supervised contrastive learning with pretext task regularization for small-scale datasets},'' no. arXiv:2108.13939, Aug. 2021, arXiv:2108.13939 [cs]. [Online]. Available: \url{http://arxiv.org/abs/2108.13939}
\BIBentrySTDinterwordspacing

\bibitem{joshi2024data}
S.~Joshi, A.~Jain, A.~Payani, and B.~Mirzasoleiman, ``Data-efficient contrastive language-image pretraining: Prioritizing data quality over quantity,'' in \emph{International Conference on Artificial Intelligence and Statistics}.\hskip 1em plus 0.5em minus 0.4em\relax PMLR, 2024, pp. 1000--1008.

\bibitem{wen2021toward}
Z.~Wen and Y.~Li, ``Toward understanding the feature learning process of self-supervised contrastive learning,'' in \emph{International Conference on Machine Learning}.\hskip 1em plus 0.5em minus 0.4em\relax PMLR, 2021, pp. 11\,112--11\,122.

\bibitem{qian2022makes}
H.~Qian, T.~Tian, and C.~Miao, ``What makes good contrastive learning on small-scale wearable-based tasks?'' in \emph{Proceedings of the 28th ACM SIGKDD conference on knowledge discovery and data mining}, 2022, pp. 3761--3771.

\bibitem{schwartz2020green}
R.~Schwartz, J.~Dodge, N.~A. Smith, and O.~Etzioni, ``Green ai,'' \emph{Communications of the ACM}, vol.~63, no.~12, pp. 54--63, 2020.

\bibitem{tripp2024measuring}
C.~E. Tripp, J.~Perr-Sauer, J.~Gafur, A.~Nag, A.~Purkayastha, S.~Zisman, and E.~A. Bensen, ``Measuring the energy consumption and efficiency of deep neural networks: An empirical analysis and design recommendations,'' \emph{arXiv preprint arXiv:2403.08151}, 2024.

\bibitem{asperti2021dissecting}
A.~Asperti, D.~Evangelista, and M.~Marzolla, ``Dissecting flops along input dimensions for greenai cost estimations,'' in \emph{International Conference on Machine Learning, Optimization, and Data Science}.\hskip 1em plus 0.5em minus 0.4em\relax Springer, 2021, pp. 86--100.

\bibitem{elia-ojcoms}
E.~Guerra, F.~Wilhelmi, M.~Miozzo, and P.~Dini, ``The cost of training machine learning models over distributed data sources,'' \emph{IEEE Open Journal of the Communications Society}, vol.~4, pp. 1111--1126, 2023.

\bibitem{huang2021towards}
W.~Huang, M.~Yi, X.~Zhao, and Z.~Jiang, ``Towards the generalization of contrastive self-supervised learning,'' \emph{arXiv preprint arXiv:2111.00743}, 2021.

\bibitem{wang2020understanding}
T.~Wang and P.~Isola, ``Understanding contrastive representation learning through alignment and uniformity on the hypersphere,'' in \emph{International conference on machine learning}.\hskip 1em plus 0.5em minus 0.4em\relax PMLR, 2020, pp. 9929--9939.

\bibitem{xu2023quantifying}
J.~Xu and H.~Liu, ``Quantifying the variability collapse of neural networks,'' in \emph{International Conference on Machine Learning}.\hskip 1em plus 0.5em minus 0.4em\relax PMLR, 2023, pp. 38\,535--38\,550.

\bibitem{li2022understanding}
A.~C. Li, A.~A. Efros, and D.~Pathak, ``Understanding collapse in non-contrastive siamese representation learning,'' in \emph{European Conference on Computer Vision}.\hskip 1em plus 0.5em minus 0.4em\relax Springer, 2022, pp. 490--505.

\bibitem{masters2021making}
D.~Masters, A.~Labatie, Z.~Eaton-Rosen, and C.~Luschi, ``Making efficientnet more efficient: Exploring batch-independent normalization, group convolutions and reduced resolution training,'' \emph{arXiv preprint arXiv:2106.03640}, 2021.

\bibitem{lee2019energy}
Y.~Lee, J.-w. Hwang, S.~Lee, Y.~Bae, and J.~Park, ``An energy and gpu-computation efficient backbone network for real-time object detection,'' in \emph{Proceedings of the IEEE/CVF conference on computer vision and pattern recognition workshops}, 2019, pp. 0--0.

\bibitem{yang2021comparative}
Y.~Yang, L.~Zhang, M.~Du, J.~Bo, H.~Liu, L.~Ren, X.~Li, and M.~J. Deen, ``A comparative analysis of eleven neural networks architectures for small datasets of lung images of covid-19 patients toward improved clinical decisions,'' \emph{Computers in Biology and Medicine}, vol. 139, p. 104887, 2021.

\end{thebibliography}

\end{document}